\documentclass{article}

\usepackage{microtype}
\usepackage{graphicx}
\usepackage{subcaption}
\usepackage{booktabs}
\usepackage{multirow}
\usepackage[table]{xcolor}
\usepackage{hyperref}

\usepackage[accepted]{icml2026}

\usepackage{amsmath}
\usepackage{amssymb}
\usepackage{mathtools}
\usepackage{amsthm}
\usepackage[capitalize,noabbrev]{cleveref}

\definecolor{methodrow}{HTML}{F1F1F1}

\theoremstyle{plain}

\theoremstyle{definition}

\theoremstyle{remark}

\icmltitlerunning{Natural-Language-Guided Binder Shortlisting}

\begin{document}

\twocolumn[
  \icmltitle{Natural-Language-Guided Generator-Agnostic Shortlisting\\for Protein Binder Design}
  % \icmlsetsymbol{equal}{*}
  \begin{icmlauthorlist}
    \icmlauthor{Gyubok Lee}{kaist}
    \icmlauthor{Kiwoong Yoo}{lgai}
    \icmlauthor{Jimin Seo}{snu}
    \icmlauthor{Jiyoun Kim}{kaist}
    \icmlauthor{Kyunghoon Hur}{keti}
    \icmlauthor{Edward Choi}{kaist}
  \end{icmlauthorlist}
    \icmlaffiliation{kaist}{Kim Jaechul Graduate School of AI, Korea Advanced Institute of Science and Technology (KAIST), Daejeon, South Korea}
    \icmlaffiliation{lgai}{LG AI Research, Seoul, South Korea}
    \icmlaffiliation{snu}{Department of Electrical and Computer Engineering, Seoul National University, Seoul, South Korea}
    \icmlaffiliation{keti}{Korea Electronics Technology Institute (KETI), Seongnam, South Korea}
\icmlcorrespondingauthor{Gyubok Lee}{gyubok.lee@kaist.ac.kr}
  \icmlkeywords{Protein Design, Large Language Models, Binder Shortlisting}
  \vskip 0.3in
]

\printAffiliationsAndNotice{}

\begin{abstract}
Modern de novo design workflows generate many candidate protein binders, but wet-lab validation capacity remains limited, making shortlisting a major bottleneck. We study whether LLMs can generate multi-metric ranking policies from precomputed structural-confidence and interface-quality proxy scores. Rather than proposing a new protein binder design pipeline, we focus on \emph{post-generation binder shortlisting}: selecting the final top-$K$ candidates from already generated binder pools using a shared panel of precomputed proxy scores. On the 10-target held-out split, averaging performance over five sampled global iterative \texttt{gpt-4o} policies reaches 0.589 $\mathrm{Recall}@10$, modestly improving over the strongest single-feature fixed baseline, Protenix binder ipTM, which reaches 0.571 $\mathrm{Recall}@10$. On the 3-target held-out subset comprising Nipah, RBX1, and TREM2, target-conditioned iterative \texttt{gpt-5.4} policies reach the strongest LLM performance, with 0.519 $\mathrm{Recall}@10$ and 0.583 $\mathrm{NDCG}@10$. These results suggest that LLM-generated ranking policies can act as an interpretable post-generation decision layer for combining heterogeneous proxy metrics to prioritize binders from large candidate pools.
\end{abstract}

\section{Introduction}
\label{sec:intro}

Recent de novo protein binder pipelines couple generative backbone models such as RFdiffusion~\cite{watson2023rfdiffusion}, ProteinMPNN-style sequence designers~\cite{dauparas2022proteinmpnn}, and structure-prediction filters based on AlphaFold2~\cite{jumper2021alphafold,evans2021multimer} or Boltz-2~\cite{boltz2}. These advances have made large-scale candidate generation increasingly routine, but experimental validation capacity remains limited. As a result, shortlisting has become a central determinant of how efficiently generated binders are converted into validated hits~\cite{bennett2023improving,pacesa2025bindcraft,proteinbase2026nipah}.

\begin{figure*}[t]
\centering
\includegraphics[width=\linewidth]{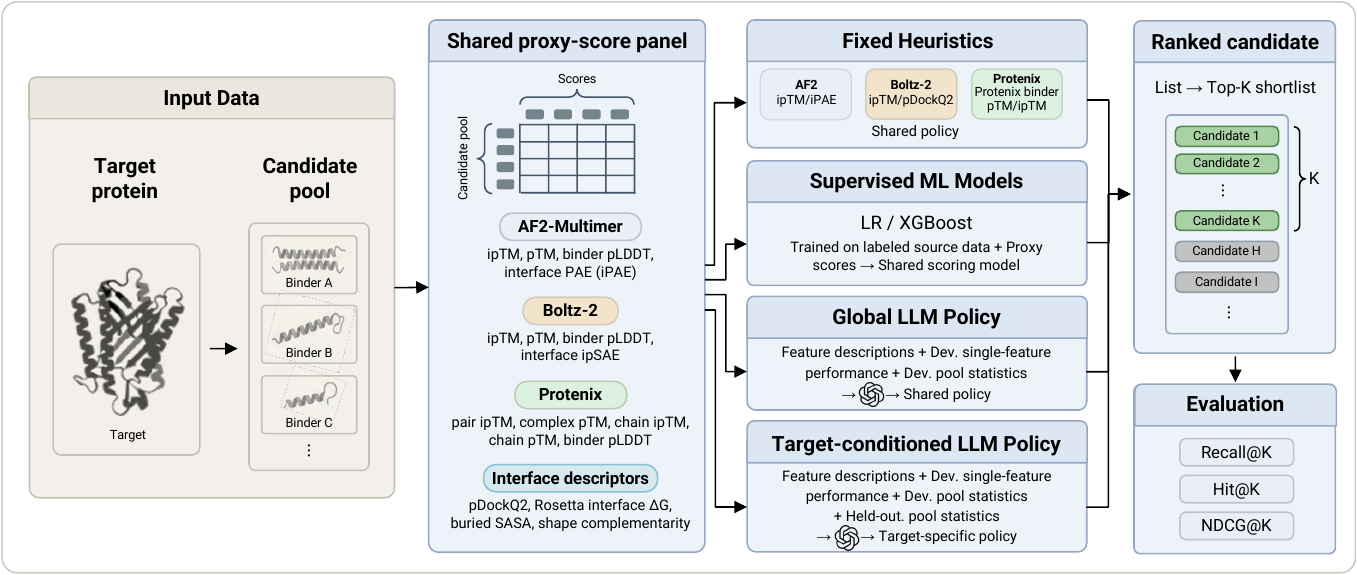}
\caption{Overview of the post-generation binder shortlisting task. For each target protein, a fixed pool of candidate binders is generated in advance by upstream design workflows. A shared proxy-score panel is then computed for all candidates using AF2-Multimer, Boltz-2, Protenix/PXDesign, and interface descriptors such as Rosetta interface $\Delta G$. Shortlisting methods, including fixed heuristics, supervised ML models, global LLM policies, and target-conditioned LLM policies, rank the candidates and return a top-$K$ shortlist for experimental testing. Global LLM policies apply a single shared ranking rule to all held-out targets, whereas target-conditioned LLM policies generate a target-specific ranking rule.}
\label{fig:overview}
\end{figure*}

A common practice is to prioritize or filter candidates using fixed thresholds, single-score rankings, or hand-tuned combinations of structure-prediction confidence scores and interface-centric proxies~\cite{bennett2023improving,pacesa2025bindcraft,proteinbase2026nipah}. Representative examples include BindCraft's fixed AF2/Rosetta filter set, Adaptyv's Boltz-2 ipSAE-based computational selection, and PXDesign's Protenix-based confidence filters~\cite{pacesa2025bindcraft,proteinbase2026nipah,bytedance2025protenix,ren2025pxdesign}. These workflow-specific filters are important for candidate curation, but they do not fully resolve the final selection problem: after designs have been generated, filtered, or collected from different workflows, only a small number can be experimentally tested. A recent meta-analysis of 3,766 experimentally tested de novo binders reports that interface-focused confidence metrics such as ipSAE and orthogonal physicochemical descriptors can improve binder selection, while predictive performance still varies substantially by target~\cite{overath2025metaanalysis}. Together, these observations motivate a post-generation shortlisting setting that combines complementary proxy scores and tests whether the ranking rule should be global or target-conditioned.

To study this setting, we separate shortlisting from candidate binder generation and treat it as a post-generation decision problem. For each target protein, we fix the generated candidate pool and a common 17-feature panel of precomputed proxy scores before shortlisting. The panel combines model-native confidence scores from AF2-Multimer, Boltz-2, and Protenix with interface-level proxy descriptors computed from predicted complexes. We use \emph{policy} broadly to denote any deterministic shortlisting rule that maps candidate proxy scores to a ranked or selected subset. We compare fixed heuristics, supervised machine learning (ML) baselines, and large language model (LLM)-based shortlisting methods that generate either a single global policy for all held-out targets or a separate target-conditioned policy for each held-out target.

Our contributions are threefold:
\begin{itemize}\itemsep0pt
\item \textbf{A post-generation shortlisting task.} We formulate binder shortlisting as final candidate selection from fixed generated pools using a common 17-feature proxy panel and a shared top-$K$ recall protocol.
\item \textbf{A controlled comparison of shortlisting strategies.} We compare fixed single-feature ranking heuristics, supervised ML baselines, global LLM policies, and target-conditioned LLM policies on the same held-out candidate pools and top-$K$ metrics.
\item \textbf{LLM-generated ranking policies provide a competitive post-generation decision layer.} Across held-out binder pools, iterative LLM policies synthesize interpretable feature-weighted combinations of structural-confidence and interface-quality proxy scores. These policies are competitive with strong single-feature and supervised baselines in top-\(K\) recall, and in the best settings modestly improve Recall@10.
\end{itemize}

\section{Related Work}
\label{sec:related}

\paragraph{Binder pipelines and candidate selection.}
Modern binder-design workflows often combine RFdiffusion backbone generation~\cite{watson2023rfdiffusion}, ProteinMPNN-style sequence design~\cite{dauparas2022proteinmpnn}, and structure-based validation or filtering with predictors such as AF2~\cite{bennett2023improving}. Existing systems typically implement candidate selection through workflow-specific filters or ranking rules. BindCraft~\cite{pacesa2025bindcraft} uses fixed AF2-confidence, Rosetta, and interface-quality filters; Adaptyv's Nipah release used Boltz-2 ipSAE ranking together with community voting and expert curation~\cite{proteinbase2026nipah}; and PXDesign~\cite{ren2025pxdesign} uses Protenix/AF2-based filtering and releases PXDesignBench for standardized monomer and binder evaluation. These pipelines show that predictor-derived confidence and interface metrics are useful for binder triage, but they leave open how to combine heterogeneous proxy scores once a fixed candidate pool must be shortlisted for experimental testing. Consistent with this gap, a 3{,}766-binder meta-analysis reports that ipSAE-based scores outperform common interface-confidence metrics and that Rosetta-derived descriptors provide complementary signal, while predictive performance remains target-dependent~\cite{overath2025metaanalysis}. Our work is complementary: we do not introduce a generator or reproduce a pipeline-specific filtering stack, but study a post-generation, generator-agnostic shortlisting layer over heterogeneous candidate pools using fixed proxy scores from multiple predictor families.

\begin{table*}[t]
\centering
\caption{Composition of the development and held-out evaluation datasets after target-level merging and exact sequence de-duplication. Counts report designs, experimentally confirmed binders, and targets for each source. For pMHC targets, the table reports short target names. The full peptide--HLA pairs are SLLMWITQC--HLA-A*02:01 and RVTDESILSY--HLA-A*01:01.}
\label{tab:datasplit}
\scriptsize
\setlength{\tabcolsep}{3pt}
\renewcommand{\arraystretch}{0.9}
  \begin{tabular}{p{0.11\textwidth}p{0.15\textwidth}p{0.24\textwidth}p{0.27\textwidth}rrr}
\toprule
Split & Source & Candidate origin & Target names & Designs & Binders & Targets \\
\midrule
Development &
BoltzGen &
BoltzGen &
AMBP, HNMT, IDI2, IL-7Ra, Insulin Receptor, MZB1/PERP1, PDGFR Beta, PD-L1, PHYH, PMVK, RFK &
327 & 103 & 11 \\
\midrule
  Held-out eval. &
BindCraft1 revalidation &
BindCraft &
IFNAR2, spCas9, Der f 21, Der f 7 &
76 & 31 & 4 \\
&
Merged EGFR &
Mixed participant-submitted methods, including Adaptyv EGFR competition submissions and BindCraft EGFR revalidation &
EGFR &
605 & 68 & 1 \\
&
pMHC minibinders &
RFdiffusion + ProteinMPNN + AlphaFold2 filtering &
NY-ESO-1 pMHC; RVTDESILSY pMHC &
137 & 3 & 2 \\
&
Nipah release &
Mixed participant-submitted methods &
Nipah Virus Glycoprotein G (NiV-G) &
1{,}030 & 103 & 1 \\
&
  GEM/Adaptyv &
  Mixed participant-submitted methods &
  RBX1 &
  321 & 9 & 1 \\
  &
  BioArena/Adaptyv &
  Mixed human/agent submissions with heterogeneous methods &
  TREM2 &
  100 & 37 & 1 \\
  \cmidrule(l){2-7}
  & \multicolumn{3}{r}{\textbf{Held-out subtotal}} &
  \textbf{2{,}269} & \textbf{251} & \textbf{10} \\
  \midrule
  \multicolumn{4}{r}{\textbf{Overall total (development + held-out)}} &
  \textbf{2{,}596} & \textbf{354} & \textbf{21} \\
\bottomrule
\end{tabular}
\renewcommand{\arraystretch}{1.0}
\end{table*}

\paragraph{Target-aware binder design.}
\citet{cao2022design} generate binders from target structure alone. \citet{gainza2023denovo} learn surface fingerprints to parameterize interaction design. APPRAISE~\cite{ding2024appraise} ranks engineered proteins by target-binding propensity through structure modeling. These works focus on generation or pairwise compatibility scoring. Our work differs in task: given a generated pool and precomputed proxy scores, we synthesize a separate decision policy for selecting $K$ candidates.

\paragraph{LLMs and agents for protein design.}
ProtAgents~\cite{ghafarollahi2024protagents} frames protein discovery as a multi-agent collaboration among LLM-backed roles that can retrieve knowledge, analyze structures, and call physics or machine-learning tools. ProteinCrow~\cite{ponnapati2025proteincrow} similarly builds an agentic protein-design assistant around curated tools, structural inputs, literature, and biochemical context. More broadly, \citet{lee2025language} review language-model use in protein design, including sequence modeling, context-conditioned design, and structure integration. Our use of LLMs is complementary: we apply them at the post-generation decision layer, where they propose ranking policies for constructing final shortlists from already-generated candidate pools.

\section{Problem Setup}
\label{sec:problem}

\paragraph{Shortlisting as policy synthesis.}
For a target protein $t$, we are given a generated candidate binder pool $\mathcal{C}_t$ of $n_t$ designs. Each candidate $c \in \mathcal{C}_t$ has a fixed proxy-score vector $\mathbf{x}_{t,c} \in \mathbb{R}^{m}$ computed before shortlisting, where $m$ is the number of common features available for every candidate. The task is to choose a subset $S_t \subseteq \mathcal{C}_t$ of size $K$ that maximizes recall of experimentally verified binders under the validation budget. A method therefore outputs a ranking policy $\pi_t \in \Pi$, and a deterministic executor scores every candidate by $\pi_t$ and returns the top $K$. In this formulation, ordinary metric-based ranking is a special case: sorting by one score, such as Boltz-2 ipSAE, is a one-feature policy. Policy synthesis generalizes this by choosing which proxy scores to combine and how strongly to weight them for the target pool.

\paragraph{Policy space.}
We use the term \emph{policy} for the structured triple $(F, w, g)$ that specifies a ranking function. Here $F \subseteq \{f_1,\dots,f_m\}$ selects a subset of features, $w \in \{1,2,3\}^{|F|}$ assigns integer weights, and each selected feature has a pre-defined higher-is-better or lower-is-better direction. In the main LLM policy space, $g$ is a weighted normalized sum: the executor normalizes every selected feature within the target pool, computes the aggregate score, sorts candidates in descending order, and returns the top $K$.

\section{Datasets}
\label{sec:data}

\subsection{Sources and split}

We collected labeled binder-design data from eight public sources or workflow releases. A source is included only when it reports candidate-level experimental outcomes for tested designs, so that each target defines a retrospective shortlisting episode: the candidate pool is fixed, and every candidate has a binder/non-binder label.

Table~\ref{tab:datasplit} summarizes the target-disjoint development and held-out splits. The \textbf{development split} contains 11 targets from BoltzGen, a de novo binder-generation workflow and validation dataset~\cite{stark2025boltzgen}. The \textbf{10-target held-out split} combines independent workflow outputs and public validation releases, including BindCraft revalidation~\cite{pacesa2025bindcraft}, pMHC minibinders, Nipah~\cite{proteinbase2026nipah}, RBX1~\cite{proteinbase2026rbx1}, TREM2~\cite{proteinbase2026trem2}, and merged EGFR challenge/revalidation pools; its experimental labels were released after the documented \texttt{gpt-4o-2024-11-20} cutoff date used for the knowledge-leakage audit (October 1, 2023). We also report a \textbf{3-target held-out subset} consisting of Nipah, RBX1, and TREM2, whose labels were released after the documented \texttt{gpt-5.4} cutoff date (August 31, 2025). When the same biological target appears in multiple releases, we merge the corresponding pools and de-duplicate exact candidate amino-acid sequences. EGFR, for example, combines Adaptyv R1, Adaptyv R2, and BindCraft1 revalidation into 605 unique designs. Full preprocessing details, including binding-outcome parsing, source-specific target assignment, zero-positive target handling, and sequence de-duplication, are provided in Appendix~\ref{app:preprocessing}.

\begin{table}[t]
\centering
\caption{Candidate-level proxy score panel (17 active features). Feature definitions, monotonic directions, and extraction details are in Appendix~\ref{app:features}.}
\label{tab:features}
\scriptsize
\setlength{\tabcolsep}{3pt}
\begin{tabular}{p{0.21\linewidth}p{0.31\linewidth}p{0.40\linewidth}}
\toprule
Family & Features & Description \\
\midrule
AF2-Multimer & ipTM, pTM, binder pLDDT, interface PAE & Model-native complex confidence, binder local confidence, and interface PAE from AF2-Multimer~\cite{evans2021multimer,mirdita2022colabfold}; lower interface PAE is better. \\
Boltz-2 & ipTM, pTM, binder pLDDT, ipSAE, pDockQ2 & Model-native confidence scores from Boltz-2~\cite{boltz2}, ipSAE-style interface confidence~\cite{dunbrack2025ipsae}, and pDockQ2~\cite{zhu2023pdockq2}, reported as metric-wise best summaries across target-MSA diffusion samples. \\
Protenix/PXDesign & pair ipTM, complex pTM, binder ipTM, binder pTM, binder pLDDT & Confidence fields from Protenix/PXDesign~\cite{bytedance2025protenix,ren2025pxdesign} covering complex, binder-chain, and binder-target confidence under our two-chain target-binder convention. \\
Rosetta interface descriptors & interface $\Delta G$, buried SASA, shape complementarity & Rosetta descriptors~\cite{stranges2013comparison,lawrence1993shape} are computed on Boltz-2 target-MSA predicted complexes, not experimental structures. \\
\bottomrule
\end{tabular}
\end{table}

\subsection{Feature extraction}
\label{sec:features}

Each candidate is represented by the 17 active proxy scores summarized in Table~\ref{tab:features}. Boltz-2 pDockQ2 and ipSAE are post-processed interface-quality or interface-confidence proxies computed from Boltz-2 predicted complexes and confidence outputs; pDockQ2 follows \citet{zhu2023pdockq2}, and ipSAE follows the PAE-based interprotein scoring approach of \citet{dunbrack2025ipsae}. Rosetta InterfaceAnalyzer metrics are computed on Boltz-2 target-side-MSA complexes and used as geometry, burial, and energy proxy scores, not ground-truth binding energies. Protenix/PXDesign features provide complex, binder-chain, and pairwise binder-target confidence; Protenix ipTM-style scores are not assumed to be numerically calibrated to AF2-Multimer or Boltz-2 ipTM. All evaluated candidates have non-missing values for the 17 active features. Appendix~\ref{app:features} gives the active features, monotonic directions, and extraction summaries.

\paragraph{Inference settings.}
Feature extraction uses target-side-MSA complex predictions where available. For each target, we precompute one target-chain MSA and reuse it for all candidate binders; the de novo binder chain is kept single-sequence because designed binders have no natural homologs. We run AF2-Multimer with 3 models and 3 recycles, and Boltz-2 with 5 diffusion samples, 3 recycling steps, and 200 sampling steps. Protenix/PXDesign predictions likewise provide the target chain with the precomputed target MSA while keeping the binder chain single-sequence. Rosetta InterfaceAnalyzer is applied to Boltz-2 target-side-MSA predicted complexes. Templates are disabled in all complex-prediction runs.

\section{Experimental Setup}
\label{sec:setup}

\subsection{Shortlisting methods}
\label{sec:methods}

We use \emph{global} to denote methods that use one rule unchanged across all held-out targets, and \emph{target-conditioned} to denote methods that generate a separate rule for each held-out target. A target-conditioned method may choose different feature subsets or weights for different candidate pools.

\paragraph{Fixed ranking heuristics.}
We evaluate target-agnostic fixed rules as single-score references spanning the main structure-prediction signals used for binder triage: AF2-Multimer ipTM and interface PAE (lower is better)~\cite{evans2021multimer,mirdita2022colabfold}, Boltz-2 ipTM and a post-processed interface-quality estimate (pDockQ2 computed from Boltz-2 predicted complexes)~\cite{boltz2,zhu2023pdockq2}, and Protenix/PXDesign binder-chain confidence (binder pTM and binder ipTM)~\cite{bytedance2025protenix,ren2025pxdesign}. Each rule ranks candidates within a target pool by one score only, testing how far a commonly used single metric can go before any learned or LLM-composed policy is introduced. The Protenix binder metrics are the binder-chain confidence fields used by the PXDesign Protenix filters, evaluated here as single-score ranking heuristics rather than hard thresholds.

\paragraph{Logistic regression and XGBoost.}
We evaluate supervised ML baselines based on logistic regression and XGBoost~\cite{chen2016xgboost}. The first fits a logistic regression model with no regularization penalty and an XGBoost model on the 11-target BoltzGen development split using the same 17-feature panel as the LLM policies; these models test whether direct supervised learning over the feature panel is sufficient without target-conditioned policy synthesis. The second is a transfer baseline using Cao binder pools~\cite{cao2022design} with retrospective AF2 scores from \citet{bennett2023improving}. For this transfer setting, we use the available AF2 interaction pAE and binder pLDDT scores as the closest historical counterparts to our AF2 interface PAE and binder-chain pLDDT features. These transfer features come from the AF2 scores previously computed by Bennett et al. for the Cao binder pools, rather than our AF2-Multimer target-side-MSA feature extraction, so they test cross-protocol transfer rather than a matched supervised re-training setting. Cao targets that overlap evaluation targets (EGFR, IL7Ra, and PDGFR) are removed before fitting. At evaluation time, each supervised model assigns every held-out candidate a fitted probability of being a binder, and candidates are ranked by this probability in descending order.

\paragraph{LLM policy sampling and averaging.}
We evaluate four LLM policy settings with \texttt{gpt-4o} on the 10-target held-out split and on the 3-target held-out subset, and we evaluate the corresponding \texttt{gpt-5.4} settings on the same 3-target held-out subset. All \texttt{gpt-4o} results use the \texttt{gpt-4o-2024-11-20} API snapshot. In the \emph{global LLM} setting, the LLM sees natural-language feature descriptions, per-development-target pool distribution summaries, and development-set single-feature performance, emits one global policy, and that policy is fixed before held-out evaluation. The distribution summaries are reported separately for each development target pool rather than pooled across candidates, while the single-feature performance values are target-averaged Recall@10/Hit@10/NDCG@10 values from ranking each development target with one feature at a time. In the \emph{global iterative LLM} setting, the final accepted policy from development-split iterative search is likewise fixed and applied unchanged to every held-out target.

The \emph{target-conditioned LLM} setting is a label-free test-time adaptation setting: it instantiates one prompt per held-out target and includes the same development calibration context as the global LLM prompt, plus that held-out target pool's identifier and unlabeled score-distribution statistics computed only within the current candidate pool. Thus, the single-turn global and target-conditioned prompts share the development-side information, while target-conditioned prompting additionally exposes unlabeled current-target context and emits one rule per target. The \emph{target-conditioned iterative LLM} setting additionally receives accepted/rejected development-search feedback from prior policy evaluations.

In all LLM settings, each sampled policy selects 3 to 5 features and assigns positive integer weights in \(\{1,2,3\}\). Each LLM policy defines a weighted rank score over selected features: selected features are normalized to a 0--1 range within the target pool, lower-is-better features are direction-corrected so that larger normalized values are better, candidates are sorted by the resulting weighted score in descending order, and the top \(K\) candidates are selected. Appendix~\ref{app:llm-prompts} summarizes the information available to each prompting setting and the shared policy constraints for the 17-feature panel. The reported LLM results average performance over five independently generated policies, which reduces run-to-run variability without treating policy averaging as a separate shortlisting method.

For proxy-score family ablations, we remove all selected terms belonging to one feature family from each sampled target-conditioned iterative policy, keep the remaining weights fixed, and re-evaluate the same deterministic executor. The LLM is not asked to regenerate or repair the policy after feature removal. Because proxy-score families are correlated and the remaining policy is not re-optimized, these ablations are interpreted as policy-dependence checks rather than monotonic feature-importance estimates. For model-to-model comparison, the main ablation table uses the shared 3-target held-out subset evaluated for both \texttt{gpt-4o} and \texttt{gpt-5.4}.

\subsection{Evaluation protocol}
\label{sec:eval}

\paragraph{Protocol.}
$K=10$ is fixed before evaluation. The primary metric is Recall@10 over verified binders, with denominator $\min(K, n_{\text{binders}})$ so that targets with fewer than 10 verified binders can still attain a maximum score of 1.0 by recovering all positives. Secondary metrics are Precision@10 and normalized discounted cumulative gain (NDCG@10). Precision@10 is the wet-lab hit rate among the 10 selected designs, whereas NDCG@10 measures whether verified binders are concentrated near the top of the shortlist; its ideal DCG is computed with the same $\min(K, n_{\text{binders}})$ number of positives for each target. Statistics are averaged across target pools rather than pooled across individual candidates.

\paragraph{Held-out evaluation.}
Target-conditioned policies use unlabeled held-out pool statistics at test time, whereas fixed heuristics, supervised baselines, and global LLM variants are fixed before held-out evaluation and applied without held-out pool statistics. Additional provenance checks, historical dataset handling, and remaining leakage caveats are provided in Appendix~\ref{app:temporal-leakage} and Section~\ref{sec:analysis}.

\section{Results}
\label{sec:results}

\begin{table*}[t]
\centering
\caption{Held-out evaluation ($K=10$), grouped by method type. Recall@10 uses denominator \(\min(K,n_{\text{binders}})\); Hit@10 is equivalent to Precision@10. Results are shown for the 10-target held-out split and for the 3-target held-out subset containing Nipah, RBX1, and TREM2. Dashes mark model/split combinations not reported. LLM results average five sampled policies.}
\label{tab:zeroshot}
\footnotesize
\setlength{\tabcolsep}{6pt}
\begin{tabular*}{\textwidth}{@{\extracolsep{\fill}}lcccccc@{}}
\toprule
& \multicolumn{3}{c}{10-target held-out split} & \multicolumn{3}{c}{3-target held-out subset} \\
\cmidrule(lr){2-4}\cmidrule(l){5-7}
Method & Recall@10 & Hit@10 & NDCG@10 & Recall@10 & Hit@10 & NDCG@10 \\
\midrule
\rowcolor{methodrow}\multicolumn{7}{@{}l}{\emph{Fixed ranking heuristics}} \\
AF2 ipTM & 0.437 & 0.360 & 0.417 & 0.433 & 0.433 & 0.487 \\
AF2 interface PAE & 0.417 & 0.290 & 0.337 & 0.300 & 0.300 & 0.298 \\
Boltz-2 ipTM & 0.403 & 0.290 & 0.377 & 0.133 & 0.133 & 0.166 \\
Boltz-2 pDockQ2 & 0.515 & 0.340 & 0.383 & 0.400 & 0.400 & 0.416 \\
Protenix binder pTM & 0.455 & 0.280 & 0.327 & 0.233 & 0.233 & 0.222 \\
Protenix binder ipTM & 0.571 & 0.360 & 0.525 & 0.504 & 0.500 & 0.506 \\
\midrule
\rowcolor{methodrow}\multicolumn{7}{@{}l}{\emph{Supervised ML baselines}} \\
LR-BG, no reg., 17 feat. & 0.537 & 0.370 & 0.438 & 0.507 & 0.500 & 0.503 \\
XGB-BG, 17 feat. & 0.425 & 0.270 & 0.390 & 0.267 & 0.267 & 0.284 \\
LR-Cao AF2 & 0.350 & 0.260 & 0.282 & 0.267 & 0.267 & 0.246 \\
XGB-Cao AF2 & 0.397 & 0.280 & 0.337 & 0.267 & 0.267 & 0.272 \\
\midrule
\rowcolor{methodrow}\multicolumn{7}{@{}l}{\emph{LLM policies averaged over five samples: \texttt{gpt-4o}}} \\
Global LLM & 0.435 & 0.310 & 0.372 & 0.400 & 0.400 & 0.431 \\
Target-conditioned LLM & 0.498 & 0.378 & 0.445 & 0.493 & 0.493 & 0.526 \\
Global iterative LLM & \textbf{0.589} & \textbf{0.404} & 0.494 & \textbf{0.504} & \textbf{0.500} & \textbf{0.529} \\
Target-conditioned iterative LLM & 0.584 & 0.394 & \textbf{0.523} & 0.497 & 0.493 & 0.520 \\
\rowcolor{methodrow}\multicolumn{7}{@{}l}{\emph{LLM policies averaged over five samples: \texttt{gpt-5.4}}} \\
Global LLM & -- & -- & -- & 0.467 & 0.467 & 0.495 \\
Target-conditioned LLM & -- & -- & -- & 0.502 & 0.500 & 0.514 \\
Global iterative LLM & -- & -- & -- & 0.470 & 0.467 & 0.513 \\
Target-conditioned iterative LLM & -- & -- & -- & \textbf{0.519} & \textbf{0.513} & \textbf{0.583} \\
\bottomrule
\end{tabular*}
\end{table*}

\paragraph{10-target held-out split.}
Table~\ref{tab:zeroshot} reports the held-out evaluation with separate columns for the 10-target held-out split and the 3-target held-out subset. On the 10-target held-out split, the strongest single-feature fixed baseline is Protenix binder ipTM, with Recall@10~=~0.571, Hit@10~=~0.360, and NDCG@10~=~0.525. Averaging five global iterative \texttt{gpt-4o} policies gives the highest LLM Recall@10 and Hit@10, reaching Recall@10~=~0.589 and Hit@10~=~0.404, while target-conditioned iterative \texttt{gpt-4o} gives the highest LLM NDCG@10 (0.523). These results support the interpretation that LLM policies provide inspectable multi-feature ranking rules that can outperform strong predictor-native single-score baselines in Recall@10, while predictor-native single-score rankings remain strong NDCG baselines.

\paragraph{3-target held-out subset.}
The 3-target held-out subset includes Nipah, RBX1, and TREM2. On this subset, target-conditioned iterative \texttt{gpt-5.4} reaches the strongest LLM performance, with Recall@10~=~0.519, Hit@10~=~0.513, and NDCG@10~=~0.583. The strongest fixed and supervised baselines are Protenix binder ipTM (0.504/0.500/0.506) and LR-BG without regularization (0.507/0.500/0.503). XGBoost remains unstable in this small setting: it performs well on TREM2 alone but selected no verified binders for either Nipah or RBX1.

\begin{figure*}[t]
\centering
\includegraphics[width=\textwidth]{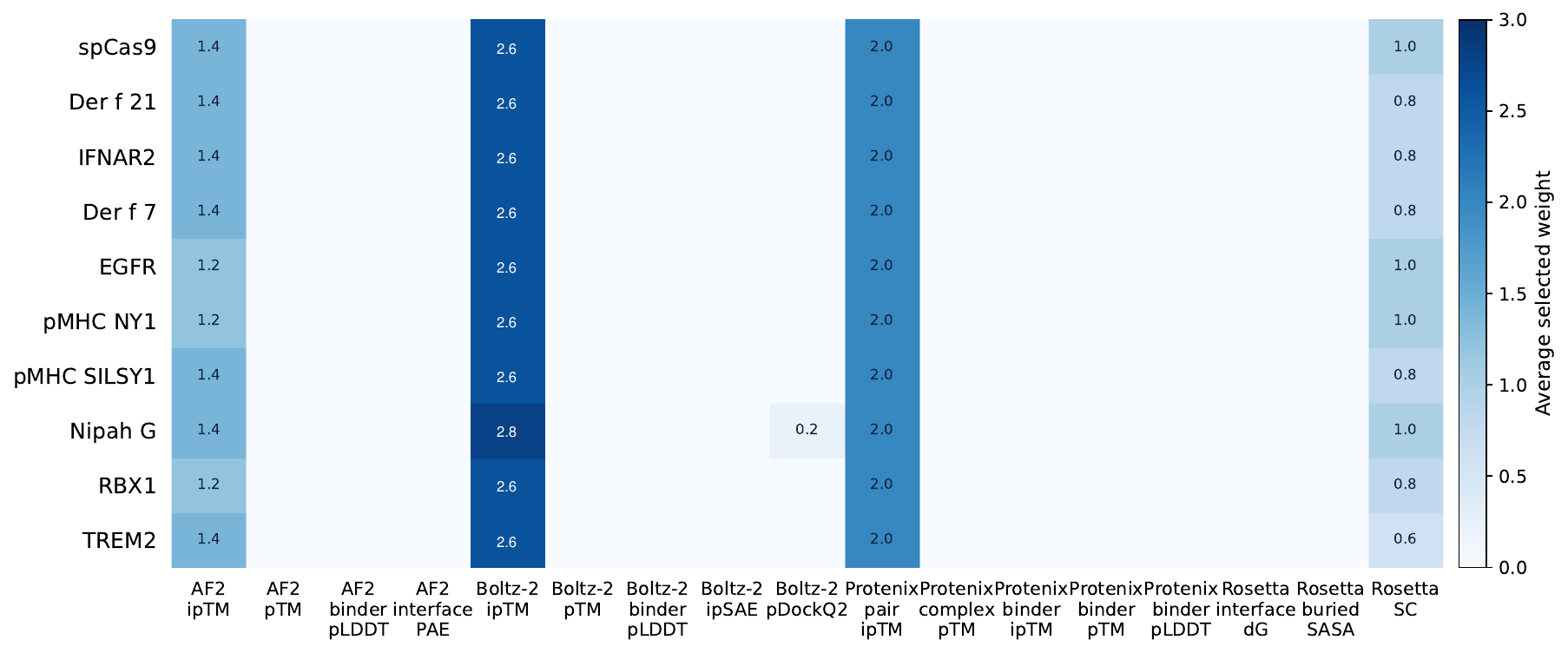}
\caption{Feature weights in target-conditioned iterative \texttt{gpt-4o} ranking policies over the full 17-feature panel. The vertical axis lists held-out targets, and the horizontal axis lists all proxy-score features. Colored cells indicate features selected by the policies; the color scale shows the average selected weight across five sampled policies. Near-zero background cells indicate features not included in the ranking policy. For lower-is-better features, positive plotted weights apply to the direction-corrected normalized score used by the executor.}
\label{fig:policy-feature-weights}
\end{figure*}

\paragraph{Generated LLM policy composition.}
The target-conditioned iterative \texttt{gpt-4o} policies concentrate weight on a small set of structure-confidence and interface-quality scores rather than spreading weight uniformly (Figure~\ref{fig:policy-feature-weights}). Across the five policy samples for each of the 10 held-out targets, every policy selects Boltz-2 ipTM and Protenix pair ipTM. Rosetta shape complementarity appears in 40 of 50 policies, AF2 ipTM appears in 27, and Boltz-2 pDockQ2 appears once. By total selected weight, the policies allocate 38.6\% to Boltz-2 features, 29.2\% to Protenix features, 19.6\% to AF2 features, and 12.6\% to Rosetta interface descriptors. Thus, the generated policies do not appear to rely on arbitrary target-specific rules. Instead, they use a compact Boltz-2/Protenix confidence backbone and make target-conditioned adjustments through AF2 ipTM and Rosetta shape-complementarity inclusion.

\paragraph{Proxy-score group ablation of generated policies.}
Table~\ref{tab:family-ablation} applies the same group leave-one-out procedure to target-conditioned iterative policies on the shared 3-target held-out subset, enabling a direct \texttt{gpt-4o} versus \texttt{gpt-5.4} comparison without changing the target set. The ablations are non-monotonic because the policy features are correlated and the remaining terms are not re-optimized after removal. For \texttt{gpt-4o}, removing Protenix causes the largest Recall@10 drop, while removing AF2 or Rosetta descriptors causes smaller drops and removing Boltz-2 slightly increases Recall@10. For \texttt{gpt-5.4}, removing Rosetta descriptors or Protenix is most harmful, consistent with its stronger use of target-specific interface-geometry and Protenix signals on the post-cutoff subset. These results should be read as policy-dependence checks, not monotonic feature-importance estimates.

\begin{table}[t]
\centering
\caption{Proxy-score group ablation of target-conditioned iterative policies on the shared 3-target held-out subset. Values are mean Recall@10 after removing all selected features from one group. No removal denotes the same ablation policy set before feature removal. Boltz-2 includes Boltz-2-derived pDockQ2; Rosetta denotes Rosetta interface descriptors.}
\label{tab:family-ablation}
\small
\setlength{\tabcolsep}{0pt}
\begin{tabular*}{\columnwidth}{@{\extracolsep{\fill}}lcc@{}}
\toprule
Removed group & \shortstack{TC iter.\\\texttt{gpt-4o}\\3-target} & \shortstack{TC iter.\\\texttt{gpt-5.4}\\3-target} \\
\midrule
No removal & 0.497 & 0.519 \\
AF2 & 0.436 & 0.499 \\
Boltz-2 & 0.517 & 0.545 \\
Protenix & 0.340 & 0.453 \\
Rosetta & 0.455 & 0.423 \\
\bottomrule
\end{tabular*}
\end{table}

\begin{table*}[t!]
\centering
\caption{Example RBX1 ranking rules. Each \(\hat{x}\) is a within-target 0 to 1 normalized feature; \(+\) rewards higher raw values and \(-\) rewards lower raw values. For LLM policies, coefficients are average policy weights across five sampled policies, with unselected features contributing weight 0 to the average; therefore, an averaged rule can include more nonzero terms than any single sampled policy. Global policies use one averaged rule for every target, whereas target-conditioned policies are averaged for RBX1 specifically. Candidates are sorted by the resulting rank score in descending order, and the top \(K\) candidates are selected.}
\label{tab:example-policies}
\footnotesize
\setlength{\tabcolsep}{8pt}
\renewcommand{\arraystretch}{1.18}
\begin{tabular*}{0.96\textwidth}{@{\extracolsep{\fill}}p{0.34\textwidth}p{0.55\textwidth}@{}}
\toprule
Policy & Rank score \\
\midrule
Fixed Protenix binder ipTM, RBX1 &
\(\displaystyle \hat{x}_{\mathrm{Protenix\ binder\ ipTM}}\) \\
\addlinespace[2pt]
Fixed Boltz-2 pDockQ2, RBX1 &
\(\displaystyle \hat{x}_{\mathrm{Boltz\ pDockQ2}}\) \\
\addlinespace[2pt]
Global iterative \texttt{gpt-4o} &
\(\displaystyle 2.2\hat{x}_{\mathrm{Boltz\ ipTM}} + 2.0\hat{x}_{\mathrm{AF2\ ipTM}} + 2.0\hat{x}_{\mathrm{Protenix\ pair\ ipTM}} + 1.0\hat{x}_{\mathrm{Rosetta\ SC}} + 0.2\hat{x}_{\mathrm{Boltz\ pDockQ2}}\) \\
\addlinespace[2pt]
Target-conditioned iterative \texttt{gpt-4o}, RBX1 &
\(\displaystyle 2.6\hat{x}_{\mathrm{Boltz\ ipTM}} + 2.0\hat{x}_{\mathrm{Protenix\ pair\ ipTM}} + 1.2\hat{x}_{\mathrm{AF2\ ipTM}} + 0.8\hat{x}_{\mathrm{Rosetta\ SC}}\) \\
\addlinespace[2pt]
Global iterative \texttt{gpt-5.4} &
\(\displaystyle 2.4\hat{x}_{\mathrm{Protenix\ pair\ ipTM}} + 1.8\hat{x}_{\mathrm{AF2\ ipTM}} + 1.2\hat{x}_{\mathrm{Boltz\ pDockQ2}} + 1.2\hat{x}_{\mathrm{Rosetta\ SC}} + 0.8\hat{x}_{\mathrm{Rosetta\ buried\ SASA}} - 0.6\hat{x}_{\mathrm{Rosetta\ interface\ dG}} + 0.4\hat{x}_{\mathrm{AF2\ pTM}} - 0.4\hat{x}_{\mathrm{AF2\ interface\ PAE}} + 0.4\hat{x}_{\mathrm{Boltz\ ipTM}} + 0.2\hat{x}_{\mathrm{Protenix\ binder\ pLDDT}} + 0.2\hat{x}_{\mathrm{Protenix\ complex\ pTM}}\) \\
\addlinespace[2pt]
Target-conditioned iterative \texttt{gpt-5.4}, RBX1 &
\(\displaystyle 2.0\hat{x}_{\mathrm{Rosetta\ buried\ SASA}} + 1.8\hat{x}_{\mathrm{Protenix\ pair\ ipTM}} + 1.6\hat{x}_{\mathrm{Rosetta\ SC}} + 1.4\hat{x}_{\mathrm{Boltz\ ipSAE}} + 0.6\hat{x}_{\mathrm{Boltz\ pDockQ2}} - 0.6\hat{x}_{\mathrm{AF2\ interface\ PAE}} + 0.6\hat{x}_{\mathrm{Boltz\ ipTM}} + 0.4\hat{x}_{\mathrm{Boltz\ binder\ pLDDT}} + 0.4\hat{x}_{\mathrm{AF2\ ipTM}} + 0.2\hat{x}_{\mathrm{AF2\ pTM}}\) \\
\bottomrule
\end{tabular*}
\renewcommand{\arraystretch}{1.0}
\end{table*}

\paragraph{Example generated policies.}
Table~\ref{tab:example-policies} shows RBX1 examples, including two fixed one-feature rules and the averaged iterative policies used by the reported LLM settings. Each sampled LLM policy is constrained to select 3 to 5 features, but an averaged example rule can contain more nonzero terms because it shows the union of features selected across five samples. The selected terms concentrate on strong predictor-native and interface-quality signals, such as Boltz-2 ipTM, Protenix pair ipTM, AF2 ipTM, Rosetta burial, and Rosetta shape complementarity, rather than introducing unrelated proxy scores. For LLM policies, coefficients are average weights across five sampled policies, with unselected features contributing 0. Positive terms reward higher raw feature values, while negative terms reward lower raw feature values.

\section{Analysis and Discussion}
\label{sec:analysis}

\paragraph{Global versus target-conditioned policies.}
The global LLM uses one rule inferred from development-target information only, then applies that rule unchanged to all held-out targets. Its Recall@10 on the 10-target held-out split is 0.435. Iterative feedback raises the global \texttt{gpt-4o} policy to 0.589 Recall@10, while target-conditioned iterative \texttt{gpt-4o} reaches a similar 0.584 Recall@10 and the highest LLM NDCG@10 on the same split (0.523). Thus, target conditioning does not uniformly dominate global policy synthesis: it can improve ranking quality without improving average top-\(K\) recovery. We interpret target conditioning as label-free adaptation to target-specific score distributions: the prompt can inspect the spread, skew, and relative behavior of proxy scores within a held-out candidate pool, then adjust which feature families to trust and how strongly to weight them, while the deterministic executor still applies the same fixed directions, within-target normalization, and top-\(K\) selection rule.

\paragraph{What iterative feedback adds.}
Iterative prompting adds development-search feedback by summarizing which feature-weighted policies were accepted or rejected on the development split and reporting their aggregate metrics. This changes the LLM's role from producing a rule from development summaries alone to producing a rule calibrated by prior policy search. On the 10-target held-out split, iterative feedback helps the global \texttt{gpt-4o} setting relative to the single global policy (0.589 vs. 0.435 Recall@10). The reflection records suggest that \texttt{gpt-4o} used this feedback mainly to converge on a compact, conservative Boltz-2/Protenix backbone; target conditioning then made only modest feature changes, which explains why target-conditioned iterative \texttt{gpt-4o} improves NDCG but does not exceed the global iterative policy in Recall@10. In contrast, \texttt{gpt-5.4} reflection records more explicitly discuss target-specific compression, spread, and redundancy in unlabeled score distributions. For RBX1, for example, the target-conditioned policies shift toward Rosetta burial/packing, Protenix pair ipTM, Boltz ipSAE, and AF2 interface PAE rather than simply reusing the global confidence backbone. This more selective adaptation is consistent with the 3-target held-out subset, where target-conditioned iterative \texttt{gpt-5.4} improves over single-turn target-conditioned \texttt{gpt-5.4} in Recall@10 (0.519 vs. 0.502) and NDCG@10 (0.583 vs. 0.514).

\paragraph{Strong single-score baselines remain important.}
Protenix binder ipTM is the strongest fixed one-feature baseline in the 17-feature panel, reaching 0.571 Recall@10 on the 10-target held-out split and 0.504 Recall@10 on the 3-target held-out subset. Boltz-2 pDockQ2 also remains competitive: it estimates interface quality from the predicted binder-target complex and confidence outputs, without using an experimental or designed reference structure. The generated policies do not replace these predictor-native signals. Instead, they combine them with complementary AF2 and Rosetta evidence; for example, all 50 target-conditioned iterative \texttt{gpt-4o} policies retain Boltz-2 ipTM and Protenix pair ipTM, while 40 include Rosetta shape complementarity and 27 include AF2 ipTM.

\section{Conclusion}
\label{sec:conclusion}

We study post-generation binder shortlisting: selecting final top-\(K\) candidates from fixed generated binder pools using precomputed structural-confidence and interface-quality proxy scores. We cast this task as policy synthesis and compare fixed heuristics, supervised baselines, and LLM-generated ranking policies. On the 10-target held-out split, averaging performance over five sampled global iterative \texttt{gpt-4o} policies modestly improves over the strongest single-feature baseline, Protenix binder ipTM, in Recall@10. The generated policies do not replace structure-prediction and interface-quality metrics. Instead, they combine AF2, Boltz-2, Rosetta interface, and occasional Protenix confidence signals into interpretable feature-weighted ranking rules that can be adjusted to each target pool. This suggests that LLM-generated ranking policies can serve as an interpretable post-generation decision layer for prioritizing binders from heterogeneous candidate pools, while strong predictor-native single-score rankings should remain explicit baselines.

\section*{Acknowledgment}
This work was supported by the Institute for Information \& communications Technology Planning \& Evaluation (IITP) grant (RS-2019-II190075) and the National Research Foundation of Korea (NRF) grant (NRF-2020H1D3A2A03100945), supported by the Korea Government (MSIT).

\bibliographystyle{icml2026}
\bibliography{references}

@article{watson2023rfdiffusion,
  title={De novo design of protein structure and function with RFdiffusion},
  author={Watson, Joseph L and Juergens, David and Bennett, Nathaniel R and Trippe, Brian L and Yim, Jason and Eisenach, Helen E and Ahern, Woody and Borst, Andrew J and Ragotte, Robert J and Milles, Lukas F and others},
  journal={Nature},
  volume={620},
  number={7976},
  pages={1089-1100},
  year={2023},
  publisher={Nature Publishing Group UK London}
}

@article{bennett2023improving,
  title={Improving de novo protein binder design with deep learning},
  author={Bennett, Nathaniel R and Coventry, Brian and Goreshnik, Inna and Huang, Buwei and Allen, Aza and Vafeados, Dionne and Peng, Ying Po and Dauparas, Justas and Baek, Minkyung and Stewart, Lance and others},
  journal={Nature Communications},
  volume={14},
  number={1},
  pages={2625},
  year={2023},
  publisher={Nature Publishing Group UK London}
}

@article{jumper2021alphafold,
  title={Highly accurate protein structure prediction with AlphaFold},
  author={Jumper, John and Evans, Richard and Pritzel, Alexander and Green, Tim and Figurnov, Michael and Ronneberger, Olaf and Tunyasuvunakool, Kathryn and Bates, Russ and {\v{Z}}{\'\i}dek, Augustin and Potapenko, Anna and others},
  journal={nature},
  volume={596},
  number={7873},
  pages={583-589},
  year={2021},
  publisher={Nature Publishing Group UK London}
}

@article{evans2021multimer,
  title={Protein complex prediction with AlphaFold-Multimer},
  author={Evans, Richard and O’neill, Michael and Pritzel, Alexander and Antropova, Natasha and Senior, Andrew and Green, Tim and {\v{Z}}{\'\i}dek, Augustin and Bates, Russ and Blackwell, Sam and Yim, Jason and others},
  journal={biorxiv},
  pages={2021-10},
  year={2021},
  publisher={Cold Spring Harbor Laboratory}
}

@article{boltz2,
  title={Boltz-2: Towards accurate and efficient binding affinity prediction},
  author={Passaro, Saro and Corso, Gabriele and Wohlwend, Jeremy and Reveiz, Mateo and Thaler, Stephan and Somnath, Vignesh Ram and Getz, Noah and Portnoi, Tally and Roy, Julien and Stark, Hannes and others},
  journal={BioRxiv},
  year={2025}
}

@article{pacesa2025bindcraft,
  title={One-shot design of functional protein binders with BindCraft},
  author={Pacesa, Martin and Nickel, Lennart and Schellhaas, Christian and Schmidt, Joseph and Pyatova, Ekaterina and Kissling, Lucas and Barendse, Patrick and Choudhury, Jagrity and Kapoor, Srajan and Alcaraz-Serna, Ana and others},
  journal={Nature},
  volume={646},
  number={8084},
  pages={483-492},
  year={2025},
  publisher={Nature Publishing Group UK London}
}

@misc{proteinbase2026nipah,
  title={Nipah Competition Results},
  author={{Adaptyv Bio}},
  year={2026},
  howpublished={Proteinbase collection},
  url={https://proteinbase.com/collections/nipah-binder-competition-results},
  note={Experimental validation results released January 21, 2026}
}

@misc{proteinbase2026rbx1,
  title={{GEM x Adaptyv}: {RBX1} Binder Design Competition},
  author={{GEM Workshop} and {Adaptyv Bio}},
  year={2026},
  howpublished={Proteinbase competition page},
  url={https://proteinbase.com/competitions/gem-adaptyv-rbx1},
  note={Experimental validation results released April 26, 2026}
}

@misc{proteinbase2026trem2,
  title={{BioArena x Adaptyv}: {TREM2} Binder Design Competition},
  author={{bioArena} and {Adaptyv Bio}},
  year={2026},
  howpublished={Proteinbase competition page},
  url={https://proteinbase.com/competitions/bioarena-adaptyv-trem2},
  note={Experimental validation results released March 28, 2026}
}

@article{overath2025metaanalysis,
  title={Predicting experimental success in de novo binder design: a meta-analysis of 3,766 experimentally characterised binders},
  author={Overath, Max D and Rygaard, Andreas SH and Jacobsen, Christian P and Brasas, Valentas and Morell, Oliver and Sormanni, Pietro and Jenkins, Timothy P},
  journal={BioRxiv},
  pages={2025-08},
  year={2025},
  publisher={Cold Spring Harbor Laboratory}
}

@inproceedings{ponnapati2025proteincrow,
  title={ProteinCrow: A Language Model Agent That Can Design Proteins},
  author={Ponnapati, Manvitha and Cox, Sam and Gordon, Cade W and Hammerling, Michael J and Narayanan, Siddharth and Laurent, Jon M and Braza, James D and Hinks, Michaela M and Skarlinski, Michael D and Rodriques, Samuel G and others},
  booktitle={ICML 2025 Generative AI and Biology (GenBio) Workshop},
  year={2025}
}

@article{cao2022design,
  title={Design of protein-binding proteins from the target structure alone},
  author={Cao, Longxing and Coventry, Brian and Goreshnik, Inna and Huang, Buwei and Sheffler, William and Park, Joon Sung and Jude, Kevin M and Markovi{\'c}, Iva and Kadam, Rameshwar U and Verschueren, Koen HG and others},
  journal={Nature},
  volume={605},
  number={7910},
  pages={551-560},
  year={2022},
  publisher={Nature Publishing Group UK London}
}

@article{gainza2023denovo,
  title={De novo design of protein interactions with learned surface fingerprints},
  author={Gainza, Pablo and Wehrle, Sarah and Van Hall-Beauvais, Alexandra and Marchand, Anthony and Scheck, Andreas and Harteveld, Zander and Buckley, Stephen and Ni, Dongchun and Tan, Shuguang and Sverrisson, Freyr and others},
  journal={Nature},
  volume={617},
  number={7959},
  pages={176-184},
  year={2023},
  publisher={Nature Publishing Group UK London}
}

@article{ding2024appraise,
  title={Fast, accurate ranking of engineered proteins by target-binding propensity using structure modeling},
  author={Ding, Xiaozhe and Chen, Xinhong and Sullivan, Erin E and Shay, Timothy F and Gradinaru, Viviana},
  journal={Molecular Therapy},
  volume={32},
  number={6},
  pages={1687-1700},
  year={2024},
  publisher={Elsevier}
}

@article{ghafarollahi2024protagents,
  title={ProtAgents: protein discovery via large language model multi-agent collaborations combining physics and machine learning},
  author={Ghafarollahi, Alireza and Buehler, Markus J},
  journal={Digital Discovery},
  volume={3},
  number={7},
  pages={1389-1409},
  year={2024},
  publisher={Royal Society of Chemistry}
}

@article{lee2025language,
  title={Language models for protein design},
  author={Lee, Jin Sub and Abdin, Osama and Kim, Philip M},
  journal={Current Opinion in Structural Biology},
  volume={92},
  pages={103027},
  year={2025},
  publisher={Elsevier}
}

@article{stranges2013comparison,
  title={A comparison of successful and failed protein interface designs highlights the challenges of designing buried hydrogen bonds},
  author={Stranges, P Benjamin and Kuhlman, Brian},
  journal={Protein Science},
  volume={22},
  number={1},
  pages={74-82},
  year={2013},
  publisher={Wiley Online Library}
}

@article{alford2017rosetta,
  title={The Rosetta all-atom energy function for macromolecular modeling and design},
  author={Alford, Rebecca F and Leaver-Fay, Andrew and Jeliazkov, Jeliazko R and O’Meara, Matthew J and DiMaio, Frank P and Park, Hahnbeom and Shapovalov, Maxim V and Renfrew, P Douglas and Mulligan, Vikram K and Kappel, Kalli and others},
  journal={Journal of chemical theory and computation},
  volume={13},
  number={6},
  pages={3031--3048},
  year={2017},
  publisher={ACS Publications}
}

@article{lin2023esmfold,
  title={Evolutionary-scale prediction of atomic-level protein structure with a language model},
  author={Lin, Zeming and Akin, Halil and Rao, Roshan and Hie, Brian and Zhu, Zhongkai and Lu, Wenting and Smetanin, Nikita and Verkuil, Robert and Kabeli, Ori and Shmueli, Yaniv and others},
  journal={Science},
  volume={379},
  number={6637},
  pages={1123-1130},
  year={2023},
  publisher={American Association for the Advancement of Science}
}

@misc{lawrence1993shape,
  title={Shape complementarity at protein/protein interfaces},
  author={Lawrence, Michael C and Colman, Peter M},
  journal={Journal of molecular biology},
  volume={234},
  number={4},
  pages={946--950},
  year={1993},
  publisher={Elsevier}
}

@article{zhu2023pdockq2,
  title={Evaluation of AlphaFold-Multimer prediction on multi-chain protein complexes},
  author={Zhu, Wensi and Shenoy, Aditi and Kundrotas, Petras and Elofsson, Arne},
  journal={Bioinformatics},
  volume={39},
  number={7},
  pages={btad424},
  year={2023},
  publisher={Oxford University Press}
}

@article{mirdita2022colabfold,
  title={ColabFold: making protein folding accessible to all},
  author={Mirdita, Milot and Sch{\"u}tze, Konstantin and Moriwaki, Yoshitaka and Heo, Lim and Ovchinnikov, Sergey and Steinegger, Martin},
  journal={Nature methods},
  volume={19},
  number={6},
  pages={679-682},
  year={2022},
  publisher={Nature Publishing Group US New York}
}

@article{dunbrack2025ipsae,
  title={R{\=e}s ipSAE loquunt: What’s wrong with AlphaFold’s ipTM score and how to fix it},
  author={Dunbrack Jr, Roland L},
  journal={bioRxiv},
  year={2025}
}

@inproceedings{chen2016xgboost,
  title={Xgboost: A scalable tree boosting system},
  author={Chen, Tianqi and Guestrin, Carlos},
  booktitle={Proceedings of the 22nd acm sigkdd international conference on knowledge discovery and data mining},
  pages={785-794},
  year={2016}
}

@article{bytedance2025protenix,
  title={Protenix-advancing structure prediction through a comprehensive AlphaFold3 reproduction},
  author={ByteDance AML AI4Science Team and Chen, Xinshi and Zhang, Yuxuan and Lu, Chan and Ma, Wenzhi and Guan, Jiaqi and Gong, Chengyue and Yang, Jincai and Zhang, Hanyu and Zhang, Ke and others},
  journal={BioRxiv},
  pages={2025-01},
  year={2025},
  publisher={Cold Spring Harbor Laboratory}
}

@article{stark2025boltzgen,
  title={Boltzgen: Toward universal binder design},
  author={Stark, Hannes and Faltings, Felix and Choi, MinGyu and Xie, Yuxin and Hur, Eunsu and O’Donnell, Timothy and Bushuiev, Anton and U{\c{c}}ar, Talip and Passaro, Saro and Mao, Weian and others},
  journal={bioRxiv},
  pages={2025-11},
  year={2025},
  publisher={Cold Spring Harbor Laboratory}
}

@article{dauparas2022proteinmpnn,
  title={Robust deep learning-based protein sequence design using ProteinMPNN},
  author={Dauparas, Justas and Anishchenko, Ivan and Bennett, Nathaniel and Bai, Hua and Ragotte, Robert J and Milles, Lukas F and Wicky, Basile IM and Courbet, Alexis and de Haas, Rob J and Bethel, Neville and others},
  journal={Science},
  volume={378},
  number={6615},
  pages={49-56},
  year={2022},
  publisher={American Association for the Advancement of Science}
}

@article{ren2025pxdesign,
  title={PXDesign: Fast, modular, and accurate de novo design of protein binders},
  author={Team, Protenix and Ren, Milong and Sun, Jinyuan and Guan, Jiaqi and Liu, Cong and Gong, Chengyue and Wang, Yuzhe and Wang, Lan and Cai, Qixu and Ma, Wenzhi and others},
  journal={bioRxiv},
  pages={2025--08},
  year={2025},
  publisher={Cold Spring Harbor Laboratory}
}

\clearpage
\appendix
% appendix.tex - \input'd by manuscript.tex after \appendix.

\section{Dataset Preprocessing}
\label{app:preprocessing}

Candidate pools are defined at the target level before feature extraction. A design is included only when the public release provides a candidate amino-acid sequence, a target sequence, and a candidate-level experimental binding outcome. Entries without an explicit binding measurement are treated as unlabeled rather than as negatives and are excluded from recall-based evaluation; this criterion removes 7 Adaptyv EGFR round-1 submissions. For ProteinBase-style releases, binding outcomes are read from the release-provided evaluation records. A design is labeled positive if any intended-target binding record is positive and negative if all intended-target binding records are negative. Expression measurements and binding-strength annotations are retained as metadata, but they do not define the binary label.

Target identifiers are assigned according to the experimental assay target reported by each source. Single-target competitions, including Adaptyv EGFR R1/R2 and Nipah, use the competition target, and off-target or control assay records are not used for the binary intended-target label. BindCraft1 revalidation candidates are assigned by their assay target in \texttt{evaluations}. For BoltzGen, PDB-like structural seed identifiers are mapped to the released biological assay target rather than used directly as target names; for example, \texttt{1g13}, \texttt{3apu}, \texttt{2a1x}, and \texttt{3qkg} correspond to GM2A, ORM2, PHYH, and AMBP, respectively.

Source-level pools are retained for audit and feature-extraction checks, including BoltzGen targets with no released positives. Targets with zero positives (GM2A, ORM2, and TNF-\(\alpha\)) are excluded from recall-based evaluation because Recall@10 is undefined when the target-level positive denominator is zero. When multiple releases contain the same biological target, the evaluation pool merges those releases and de-duplicates exact candidate amino-acid sequences. If duplicate sequences have discordant labels, the merged label is positive if any duplicate record is positive. For EGFR, this merge combines Adaptyv R1, Adaptyv R2, and BindCraft1 revalidation into 605 unique candidate sequences from 615 labeled candidate records, with 68 positives after positive-if-any label aggregation.

\section{Temporal Leakage Audit}
\label{app:temporal-leakage}

Table~\ref{tab:model-cutoffs} records the documented model cutoffs used for the temporal-leakage audit, and Table~\ref{tab:temporal-leakage} records the public-release dates used for the dataset-level argument. Development entries are included for auditability because their labels are used in supervised fitting, global-policy construction, or iterative policy-search feedback; they are not held-out evaluation labels. For \texttt{gpt-4o-2024-11-20}, the documented cutoff is 2023-10-01, so the 10-target held-out split and the 3-target held-out subset both use labels released after the cutoff. The \texttt{gpt-5.4} model has a later 2025-08-31 cutoff; Nipah, RBX1, and TREM2 are the held-out targets in Table~\ref{tab:zeroshot} whose source pools and experimental labels became public after this later date. ProteinBase competition pages sometimes retain stale stage-detail text after release, so we use the overview ``Results released'' date when available. BindCraft1 revalidation is public on ProteinBase; collection-level asset timestamps are consistent with 2025-10-01, but we use the 2025-10-06 ProteinBase launch as the conservative public web-availability date. BoltzGen's manuscript was posted on bioRxiv on 2025-11-24, but validated-label summary files were already present in the Hugging Face \texttt{boltzgen/adaptyv\_data1} upload on 2025-10-27, with a stable re-upload on 2025-10-31. The BoltzGen labels are explicitly supplied only as development information rather than used as held-out outcomes.

\begin{table}[h]
\centering
\scriptsize
\setlength{\tabcolsep}{3.5pt}
\resizebox{\columnwidth}{!}{%
\begin{tabular}{p{0.34\linewidth}p{0.20\linewidth}p{0.42\linewidth}}
\toprule
Model / run & Documented cutoff & Use in paper \\
\midrule
\texttt{gpt-4o-2024-11-20} & 2023-10-01 & Reported on the 10-target held-out split and the 3-target held-out subset \\
\texttt{gpt-5.4} & 2025-08-31 & Reported on the 3-target held-out subset: Nipah, RBX1, and TREM2 \\
\bottomrule
\end{tabular}
}
\caption{Model cutoff dates used for the temporal-leakage audit.}
\label{tab:model-cutoffs}
\end{table}

\begin{table}[h]
\centering
\scriptsize
\setlength{\tabcolsep}{2.5pt}
\resizebox{\columnwidth}{!}{%
\begin{tabular}{p{0.30\linewidth}p{0.24\linewidth}p{0.18\linewidth}cc}
\toprule
Source / pool & Targets used & Public date & After \texttt{gpt-4o}? & After \texttt{gpt-5.4}? \\
\midrule
BindCraft1 revalidation & spCas9, Der f 21, IFNAR2, Der f 7, EGFR entries & 2025-10-06 & yes & yes \\
Adaptyv EGFR R1 & EGFR & 2024-10-18 & yes & no \\
pMHC minibinders & NY-ESO-1 pMHC, RVTDESILSY pMHC & 2024-12-03 & yes & no \\
Adaptyv EGFR R2 & EGFR & 2025-01-15 & yes & no \\
BoltzGen validated-label dataset & development targets only & 2025-10-27 & yes & yes \\
GEM/Adaptyv RBX1 & RBX1 & 2026-04-26 & yes & yes \\
Nipah release & Nipah G & 2026-01-21 & yes & yes \\
BioArena/Adaptyv TREM2 & TREM2 & 2026-03-28 & yes & yes \\
\midrule
Cao/Bennett historical & transfer baseline only & 2022 to 2023 & no & no \\
\bottomrule
\end{tabular}
}
\caption{Dataset-level temporal-leakage audit. Dates report conservative public availability of candidate-level experimental labels, or stable public data-package availability where explicitly noted. Development and transfer-baseline entries are shown for provenance but are not held-out evaluation labels. For the \texttt{gpt-5.4} comparison, Nipah, TREM2, and RBX1 are the held-out targets whose source pools and experimental labels became public after the documented cutoff.}
\label{tab:temporal-leakage}
\end{table}

\section{Feature Glossary}
\label{app:features}

The main policy panel contains 17 reproducible candidate-level features, listed with monotonic directions and extraction summaries in Table~\ref{tab:feature-columns}.

\begin{table}[h]
\centering
\scriptsize
\setlength{\tabcolsep}{2.5pt}
\begin{tabular}{@{}p{0.31\linewidth}p{0.12\linewidth}p{0.49\linewidth}@{}}
\toprule
Feature & Direction & Extraction summary \\
\midrule
AF2-Multimer ipTM & higher & Mean AF2-Multimer ipTM across aggregated target-MSA complex models. \\
AF2-Multimer pTM & higher & Mean AF2-Multimer pTM across aggregated target-MSA complex models. \\
AF2 binder pLDDT & higher & Mean binder-chain AF2-Multimer pLDDT across aggregated target-MSA complex models. \\
AF2 interface PAE & lower & Mean cross-chain AF2-Multimer PAE over target-binder interface residue pairs, averaged across models. \\
Boltz-2 ipTM & higher & Best Boltz-2 ipTM summary across target-MSA diffusion samples. \\
Boltz-2 pTM & higher & Best Boltz-2 pTM summary across target-MSA diffusion samples. \\
Boltz-2 binder pLDDT & higher & Best binder-chain pLDDT summary across Boltz-2 target-MSA diffusion samples. \\
Boltz-2 ipSAE & higher & Best conservative ipSAE-style interface-confidence summary across Boltz-2 target-MSA diffusion samples. \\
Boltz-2 pDockQ2 & higher & Best pDockQ2 summary across Boltz-2 target-MSA diffusion samples. \\
Protenix pair ipTM & higher & Protenix/PXDesign pairwise binder-target confidence under the two-chain target-binder convention. \\
Protenix complex pTM & higher & Protenix/PXDesign complex-level pTM for the predicted target-binder complex. \\
Protenix binder ipTM & higher & Protenix/PXDesign binder-chain ipTM under the two-chain target-binder convention. \\
Protenix binder pTM & higher & Protenix/PXDesign binder-chain pTM under the two-chain target-binder convention. \\
Protenix binder pLDDT & higher & Protenix/PXDesign binder-chain pLDDT under the two-chain target-binder convention. \\
Rosetta interface $\Delta G$ & lower & Rosetta InterfaceAnalyzer interface $\Delta G$ on the Boltz-2 target-MSA predicted complex selected for Rosetta analysis. \\
Rosetta buried SASA & higher & Rosetta buried solvent-accessible surface area on the Boltz-2 target-MSA predicted complex selected for Rosetta analysis. \\
Rosetta shape complementarity & higher & Rosetta InterfaceAnalyzer shape complementarity~\cite{lawrence1993shape} on the Boltz-2 target-MSA predicted complex selected for Rosetta analysis. \\
\bottomrule
\end{tabular}
\caption{Active features in the 17-feature policy panel. Directions indicate the monotonic orientation used by the deterministic executor. Target-MSA complex predictions use the target-chain MSA while designed binder chains remain single-sequence.}
\label{tab:feature-columns}
\end{table}

\paragraph{Inference settings.}
AF2-Multimer and AF2-monomer runs use ColabFold v1.5.5 with three models, three recycles, and no custom templates. The single-sequence complex wrapper uses single-sequence MSA mode. The target-MSA complex wrapper supplies a precomputed multimer A3M for the target chain, while the binder chain remains single-sequence because no homologs exist for a de novo binder. Target-chain A3M files are generated once with the Protenix/PXDesign-compatible MMseqs2 MSA service and cached before feature extraction.

Boltz-2 runs use 5 diffusion samples, 3 recycling steps, 200 sampling steps, and the \texttt{boltz2} model, matching BoltzGen's de novo binder refolding configuration~\cite{stark2025boltzgen}. In the single-sequence baseline, both chains are modeled without MSA information. In the target-MSA variant, the target chain receives the precomputed target MSA and the binder chain remains single-sequence.

Protenix/PXDesign features follow the public PXDesign Protenix inference setting with one sample, two diffusion steps, four recycles, templates disabled, and MSA enabled. The target chain receives the precomputed target MSA and the binder chain remains single-sequence. Complexes are represented as target-binder pairs, and binder-chain confidence fields are extracted according to the PXDesign two-chain convention. The active Protenix/PXDesign policy features are pairwise binder-target ipTM, complex pTM, binder-chain ipTM, binder-chain pTM, and binder-chain pLDDT.

The main 17-feature panel uses target-MSA Boltz-2, AF2-Multimer, Protenix, and Rosetta complex outputs. Single-sequence complex predictions are retained for ablation but are not part of the main panel. ESMFold~\cite{lin2023esmfold}, using the HuggingFace \texttt{facebook/esmfold\_v1} weights, is deterministic and run once per binder. Rosetta \texttt{InterfaceAnalyzerMover}~\cite{stranges2013comparison} with the \texttt{ref2015} scorefunction~\cite{alford2017rosetta} is applied to Boltz-2 predicted complexes after coordinate-constrained \texttt{FastRelax} pre-relaxation for side-chain packing and local relaxation.

\section{Cao/Bennett AF2 Transfer Baseline}
\label{app:cao-transfer}

The historical Cao/Bennett transfer baseline is trained on the AF2 confidence measurements that have compatible counterparts in the current held-out pools: binder-target interface PAE and binder-chain pLDDT. On the historical training pools, these are the AF2 interaction PAE and binder pLDDT scores reported with the Cao/Bennett retrospective scoring data. On the current held-out pools, we recompute the analogous quantities using our AF2-Multimer target-MSA protocol: mean cross-chain PAE over target-binder interface residue pairs and mean binder-chain pLDDT, each averaged across AF2-Multimer model runs. RMSD-based historical features are excluded because the current held-out candidates do not generally include the original designed-complex reference structures needed to compute the same designed-versus-predicted RMSD terms. Thus, this baseline tests transfer across compatible AF2 confidence feature types, not an identically matched AF2 scoring protocol.

\section{Prompt Information Conditions}
\label{app:llm-prompts}

This appendix records the prompt information conditions used to audit label exposure and distinguish global from target-conditioned policies. All settings use the same feature panel, policy class, and deterministic executor described in Section~\ref{sec:methods}; held-out labels are never included in any prompt. Table~\ref{tab:prompt-contrast} contrasts the single-turn global and target-conditioned settings, and iterative variants add aggregate development-search feedback without changing the held-out-label restriction.

\begin{table}[h]
\centering
\scriptsize
\setlength{\tabcolsep}{3pt}
\begin{tabular}{@{}p{0.27\linewidth}p{0.31\linewidth}p{0.31\linewidth}@{}}
\toprule
Prompt aspect & Global LLM & Target-conditioned LLM \\
\midrule
Policy emitted & One policy shared by all held-out targets & One policy generated separately for each held-out target \\
Feature panel & Same 17 feature descriptions and fixed directions & Same 17 feature descriptions and fixed directions \\
Development information & Per-development-target score distributions and target-averaged single-feature performance & Same development calibration context as the global prompt \\
Held-out information & No held-out target identifiers, source context, score distributions, or labels & Current held-out target identifier and unlabeled score distributions for that target pool \\
Labels available to prompt & Development labels only through aggregate single-feature performance & Development labels only through aggregate single-feature performance; no held-out labels \\
Executor & Same deterministic executor: fixed directions, within-target normalization, weighted sum, top 10 & Same deterministic executor: fixed directions, within-target normalization, weighted sum, top 10 \\
\bottomrule
\end{tabular}
\caption{Information contrast between global and target-conditioned single-turn LLM policies. Iterative variants keep the same held-out-label restriction and executor, but add aggregate development-search feedback.}
\label{tab:prompt-contrast}
\end{table}

\paragraph{Prompt inputs and constraints.}
Rather than relying on free-form natural-language recommendations, each prompt asks the model to emit a structured ranking policy. The common prompt inputs are: (i) natural-language feature descriptions and fixed monotonic directions for the 17 active proxy scores, (ii) development-set single-feature performance summarized as target-averaged Recall@10, Hit@10, and NDCG@10, and (iii) per-development-target score-distribution summaries. Target-conditioned prompts additionally include the current held-out target identifier and unlabeled score-distribution summaries for that held-out target pool. Global prompts omit held-out target identifiers and held-out score distributions, and ask for one policy that is later applied unchanged to every held-out target.

All prompts enforce the same policy constraints. A valid policy selects 3 to 5 features, assigns each selected feature an integer weight in \(\{1,2,3\}\), and returns only the structured policy object. The prompt explicitly disallows hard thresholds, feature-specific filters, treating any proxy as a direct affinity measurement, or overriding feature directions and aggregation. The deterministic executor, not the LLM, applies the fixed feature directions, normalizes selected features within the target pool, computes the weighted normalized sum, sorts candidates in descending order, and selects the top 10 designs.

\paragraph{Iterative feedback.}
Iterative settings use the same information restrictions as their single-turn counterparts. They add only aggregate development-search feedback from previously evaluated policies, including accepted or rejected feature-weight combinations and development-set summary metrics. This feedback is computed on the development split and does not include held-out labels.

\paragraph{Representative output schema.}
The saved runs used a JSON-like structured output with one list of selected terms:
\begin{quote}
\scriptsize
\begin{verbatim}
{
  "score_terms": [
    {"feature": "<feature_name>", "weight": 1|2|3}
  ],
  "rationale": "<brief policy rationale>"
}
\end{verbatim}
\end{quote}

\section{Per-target Performance}
\label{app:pertarget}

Table~\ref{tab:pertarget-gpt4o} reports per-target Recall@10 for the strongest single-feature fixed heuristic and the matched-prompt \texttt{gpt-4o} policy settings. Table~\ref{tab:pertarget-gpt54} reports the corresponding \texttt{gpt-5.4} detailed metrics only on the 3-target held-out subset used for \texttt{gpt-5.4} evaluation.

\begin{table}[h]
\centering
\scriptsize
\setlength{\tabcolsep}{3pt}
\resizebox{\columnwidth}{!}{%
\begin{tabular}{@{}lrrccccc@{}}
\toprule
Target & $n$ & $n_b$ & Fixed & Global & TC & Iter. global & Iter. TC \\
\midrule
spCas9 & 20 & 14 & 0.700 & 0.800 & 0.900 & 0.860 & 0.840 \\
Der f 21 & 16 & 4 & 0.500 & 0.250 & 0.500 & 0.550 & 0.600 \\
Nipah G & 1030 & 103 & 0.700 & 0.600 & 0.800 & 0.760 & 0.720 \\
IFNAR2 & 20 & 8 & 0.500 & 0.500 & 0.500 & 0.725 & 0.725 \\
NY-ESO-1 pMHC & 41 & 1 & 1.000 & 1.000 & 0.200 & 1.000 & 1.000 \\
Der f 7 & 20 & 5 & 0.800 & 0.200 & 0.600 & 0.760 & 0.800 \\
EGFR & 605 & 68 & 0.200 & 0.400 & 0.400 & 0.380 & 0.280 \\
RVTDESILSY pMHC & 96 & 2 & 0.500 & 0.000 & 0.400 & 0.100 & 0.100 \\
RBX1 & 321 & 9 & 0.111 & 0.000 & 0.000 & 0.111 & 0.111 \\
TREM2 & 100 & 37 & 0.700 & 0.600 & 0.680 & 0.640 & 0.660 \\
\midrule
mean & & & 0.571 & 0.435 & 0.498 & \textbf{0.589} & 0.584 \\
\bottomrule
\end{tabular}
}
\caption{Per-target Recall@10 on the 10-target held-out split for \texttt{gpt-4o}. Fixed is Protenix binder ipTM, the strongest single-feature fixed baseline in Table~\ref{tab:zeroshot}. Global and TC denote single-turn global and target-conditioned matched-prompt policies; Iter. global and Iter. TC denote the corresponding iterative policies with development-feedback memory. LLM entries are sampled-policy averages where applicable.}
\label{tab:pertarget-gpt4o}
\end{table}

\begin{table}[h]
\centering
\scriptsize
\setlength{\tabcolsep}{3pt}
\resizebox{\columnwidth}{!}{%
\begin{tabular}{@{}llrrrr@{}}
\toprule
Metric & Target & Global & TC & Iter. global & Iter. TC \\
\midrule
Recall@10 & Nipah G & 0.800 & 0.850 & 0.580 & 0.640 \\
Recall@10 & RBX1 & 0.000 & 0.056 & 0.111 & 0.156 \\
Recall@10 & TREM2 & 0.600 & 0.600 & 0.720 & 0.760 \\
Recall@10 & mean & 0.467 & 0.502 & 0.470 & \textbf{0.519} \\
\midrule
Hit@10 & Nipah G & 0.800 & 0.850 & 0.580 & 0.640 \\
Hit@10 & RBX1 & 0.000 & 0.050 & 0.100 & 0.140 \\
Hit@10 & TREM2 & 0.600 & 0.600 & 0.720 & 0.760 \\
Hit@10 & mean & 0.467 & 0.500 & 0.467 & \textbf{0.513} \\
\midrule
NDCG@10 & Nipah G & 0.849 & 0.869 & 0.620 & 0.734 \\
NDCG@10 & RBX1 & 0.000 & 0.037 & 0.141 & 0.222 \\
NDCG@10 & TREM2 & 0.637 & 0.635 & 0.777 & 0.792 \\
NDCG@10 & mean & 0.495 & 0.514 & 0.513 & \textbf{0.583} \\
\bottomrule
\end{tabular}
}
\caption{Per-target \texttt{gpt-5.4} performance on the 3-target held-out subset. Global and TC denote single-turn global and target-conditioned matched-prompt policies; Iter. global and Iter. TC denote the corresponding iterative policies with development-feedback memory. Values are reported only for Nipah, RBX1, and TREM2, the post-cutoff subset used for \texttt{gpt-5.4} evaluation.}
\label{tab:pertarget-gpt54}
\end{table}

\end{document}